# Advances in Prediction of Readmission Rates Using Long Term Short Term Memory Networks on Healthcare Insurance Data

Shuja Khalid, MSc[1]*, Francisco Matos, MD[2]*, Ayman Abunimer, MD[2], Joel Bartlett, BS[3], Richard Duszak, MD[2], Michal Horny, PhD[2], Judy Gichoya, MD[2], Imon Banerjee, PhD[4], Hari Trivedi, MD[2]

**Affiliations:**
1. University of Toronto, Department of Computer Science, 27 King's College Cir, Toronto, Canada
2. Emory University, Department of Radiology, 1364 E Clifton Rd NE, Atlanta, GA 30322
3. Georgia Institute of Technology, School of Computer Science, North Avenue, Atlanta, GA 30332
4. University of Arizona, Department of Computer Science, 1040 4th St, Tucson, AZ 85721
* Co-first authors

**Corresponding Author:**
Francisco Matos, MD
Emory University, Department of Radiology
1364 E Clifton Rd NE,
Atlanta, GA 30322
+1-404-409-5045
FMATOSQ@emory.edu

## Abstract

**Background**
30-day hospital readmission is a long-standing medical problem that affects patients' morbidity and mortality and costs billions of dollars annually. Recently, machine learning models have been created to predict risk of inpatient readmission for patients with specific diseases, however no model exists to predict this risk across all patients.

**Methods**
We developed a bi-directional Long Short Term Memory (LSTM) Network that is able to use readily available insurance data (inpatient visits, outpatient visits, and drug prescriptions) to predict 30-day readmission for any admitted patient, regardless of reason. We compare model performance with and without inclusion of 30-days of historical data prior to the initial admission, and with 7, 14, or 21 days of additional information following discharge. Finally, we added an attention layer to the model to investigate what features were most responsible for prediction in three categories: diagnosis related group, primary diagnosis, and drug therapy class.

**Results**
The top performing model achieved an ROC AUC of $0.763 \pm 0.011$ when using historical, inpatient, and post-discharge data. The LSTM model significantly outperformed a baseline random forest classifier, indicating that understanding the sequence of events is important for model prediction. Incorporation of 30-days of historical data also significantly improved model performance compared to inpatient data alone, indicating that a patient's clinical history prior to admission, including outpatient visits and pharmacy data is a strong contributor to re-admission.

**Conclusion**
Our results demonstrate that a machine learning model is able to predict risk of inpatient readmission with reasonable accuracy for all patients using structured insurance billing data. Because billing data or equivalent surrogates can be extracted from sites, such a model could be deployed to identify patients at risk for readmission before they are discharged, or to assign more robust follow up (closer follow up, home health, mailed medications) to at-risk patients after discharge.

**Introduction**

In the mid-1980's hospital readmission rates were over 20%[1]. Since then, hospital readmission rates have barely decreased by less than 4% by 2015, despite advances in technology, increased health care spending, and policy changes/incentives for value-based care to replace fee based payment[2]. According to the Healthcare Cost and Utilization Project, by 2016 the average cost for overall readmissions was $14,400, even higher than the average overall cost of the index admission ($12,500)[3]. Readmissions cost Medicare up to $26 billion annually, and $17 billion of that is estimated to be potentially avoidable[4]. From the patient perspective, re-admissions expose patients to risks of hospital acquired infections, medical errors, and emotional deconditioning.

The Centers for Medicare and Medicaid Services (CMS) defines readmission as an admission to an acute care hospital within 30 days of discharge from the same or another acute care hospital for any reason[5]. In 2012, Medicare started the Hospital Readmission Reduction Program (HRRP) in an effort to reduce expenses from unplanned 30-day readmissions and improve quality of care across the US[6]. The program assesses hospital performance by calculating an excess readmission ratio (ratio of the predicted readmission rate to the expected readmission rate) and applies a payment reduction penalty to hospitals with worse performance[5]. These payment reduction penalties impact a large proportion of hospitals nationwide. In 2020, of the 3,080 eligible hospitals, 83% received a penalty that averaged a 0.69% reduction in Medicare payments. Together these penalties amount to approximately $553 million total.[7]

Factors surrounding the drivers of readmissions are well documented and there are numerous models developed by health care administrators to predict which patients are most vulnerable. Hospitals have relied on different risk stratification scores such as LACE[8] and HOSPITAL[9,10] but have also made significant investments to develop more effective tools for assessing the risk of readmission. More recent modeling attempts have relied on Artificial Intelligence (AI)[11], however the majority of these focus on single pathologies or diseases, limiting their generalizability. Goto *et al.* predict readmission in patients with COPD in Japan achieving an overall AUC of 0.61[12]. Golas *et al.* used structured and unstructured EHR data to predict heart failure re-admissions with an AUC of 0.70[13]. Darabi *et al.* predict re-admissions in patients with ischemic stroke using post-discharge data and achieve an AUC of 0.74[14]. However there is no prior work that attempts to predict likelihood of readmission for *all* patients using readily available and structured data that does not need to be individual extracted and curated from the electronic health record (EHR).

Beyond the challenge of model development, barriers remain to the widespread adoption of risk prediction models, including "blackbox" models which cannot be interrogated, poor or unstudied model generalizability, high data extraction and curation costs for training and implementation, and inability to translate predictions into concrete actions or interventions[15–18]. These challenges must be addressed to foster adoption of ML-based clinical decision support tools that could lower their 30 day readmission rates.

In this work, we leverage the IBM Marketscan Database [19] which contains structured information derived from public and private healthcare insurance providers including inpatient and outpatient diagnostic codes, procedures, medications, and demographics[19]. This data is used to train a bi-directional Long Short-Term Memory (LSTM) network to predict 30-day risk of re-admission. LSTMs are a type of recurrent neural network (RNN) that use feedback connections to allow processing of temporal data. This architecture has been used for many long sequence processing applications, such as handwriting recognition[20] and speech recognition[21]. It is therefore well suited for long temporal encounter data processing and can account for the *sequence* of events rather than solely the presence of absence of individual features.

This work is unique in that we do not pre-select any patient cohort or disease condition, do not require any information extracted from the EHR, and are able interrogate the model using an attention mechanism that highlights features that contribute to model performance. The resultant model and weights for this work will be publicly released.

**Methods**

*Dataset*

This retrospective study is waived from IRB review due to use of the publicly available IBM Marketscan dataset[19]. We included data from the IBM Marketscan database from 2017-2019. The dataset consists of a total of several hundred continuous and discrete variables which were narrowed 98 relevant variables shown in Appendix A, and a sample of which is shown in Table 1.

*Pre-processing*

Due to the sparsity and temporal nature of the data, we first developed heuristics for dealing with missing data on a feature-by-feature basis. For discrete features, the data was one-hot-encoded[22]. For missing data, we determined that encoding "unavailability" into the dataset was more meaningful than using simple imputation techniques such as *min*, *max* or *mode* strategies. For continuous features, the missing data was replaced with –1 as opposed to using *min*, *max* or *mean* strategies to code the data unavailability in the model.

*Sequence and Outcome Generation*

Because the data is captured at irregular timesteps, the temporality was defined by *the* order in which the events occur, without considering the difference in time intervals. Figure 1 illustrates the data extraction procedure as well as class assignment. Each data sequence begins with identification of an inpatient admission and discharge. Positive samples are those in which a repeat inpatient admission occurs within 30 days of hospital discharge, and a negative sample occurs when a patient is not readmitted within 30 days (this includes patients who were never re-admitted or re-admitted after 30 days). Any hospitalization discharge occurring within 30 days of the end of the dataset time period (12/31/2018) was excluded since readmission status was unknown.

Multiple time windows were considered to evaluate their effect on model performance, and one bi-directional LSTM model (described below) was trained using each time sequence. i) Inpatient data only (*IP*), ii) Inpatient + 30 days historical data (*HIP*), and iii) Inpatient + 30 days historical + 7, 14, or 21 days of context following discharge (*CHIP$_7$*, *CHIP$_{14}$*, and *CHIP$_{21}$*, respectively). All timespans were capped at 50 samples from most recent to oldest to reduce computational load. Sequences were padded along the time axis with zeros if the existing datapoints were insufficient. Data from inpatient, outpatient, and pharmacy visits was included, when present. Because a patient might have multiple readmission events, we consider each such event independently.

The data was divided in 80% training and validation and 20% test sets. Samples were stratified by patient such the same patient did not overlap between training, validation, and test sets. As expected, negative sequences were much more prevalent in the data, so these samples were under-sampled during training such that the number of positive and negative samples was equal.

*Baseline Model*
To establish baseline performance and demonstrate that our proposed model is warranted, a random forest classifier was created using up to 50 data elements of inpatient and historical data. The choice of model was informed by the random forest classifier's ability to capture temporal insights in structured data due to its branching conditional structure[23]. The data was pre-processed by collapsing the data along the temporal axis. Each resulting sample is thus represented by a vector of length N, where N is the product of the number of features and the number of maximum timesteps (50). For the random forest model the following hyperparameters were used: estimators (100) and depth (50).

*Bi-directional LSTM Model*

We designed a stacked bi-directional LSTM[21] model for capturing temporal relations in the data (Figure 2). During experimentation, various other RNN models (single-layered gated recurrent unit GRU[24] and unidirectional LSTM [25]) were used in place of the stacked bi-directional LSTM architecture, all of which performed worse in comparison.

Our model consists of 2 stacked LSTM layers with 150 neurons, with these values being determined empirically using an exhaustive grid-search[26]. The first LSTM layer is deployed in a many-to-many configuration[27], where the outputs are subsequently fed into the second LSTM layer. The second layer is designed as many-to-one configuration and final output is passed through a batch normalization layer followed by a dropout[28] layer to avoid overfitting to the data and increasing the likelihood of faster convergence[29]. A small learning rate (0.00001) was selected to stabilize training. The specific parameters used for training are presented in Appendix B.

To understand what features contributed most to model performance, we added an attention layer to our model. Attention mechanisms are a relatively recent advancement in machine learning and have been shown to provide powerful contextualization, leading to state-of-the-art performance in various text-based tasks such as machine translation[30], biological sequence analysis[31] and named entity recognition[32]. We use this powerful method to allows the network to focus on specific aspects of a complex temporal input and improve the explainability of our proposed model. By training our model including the attention mechanism in an end-to-end fashion, we were able to probe the model for insights related to its predictions.

All model development was done using the standard Keras[33] implementation in Python[34] with statistical libraries such as Scikit-learn[35], NumPy[36] and SciPy[37]. The use of the statistical libraries was primarily to aggregate the results and capture the results presented herein. The model architecture, including the attention mechanism and its various iterations were all developed using Keras. Th attention mechanism designed for this particular task can be thought of as a separate model that is added at the end of a bi-LSTM network. This model determines which of the input features contribute most to the final prediction. The two models were concatenated to create the overall architecture illustrated in Figure 2.

**Data Availability**: The IBM Marketscan dataset is a publicly available, commercial dataset available for licensing directly through IBM. The authors are unable to release any portion of the dataset due to the terms of use.

**Results**

A total of number of 38798 unique patients were included in the analysis, where positive and negative samples were evenly sampled. There was no significant difference in mean age or gender distribution for patients who were re-admitted in 30 days compared to those who were not (Table 2), with mean patient age of 47 ± 17 years and 47% female patients. (Table 2). Race information is not provided in the MarketScan database.

Baseline model performance using the random forest classifier was poor with an ROC AUC of 0.510 ± 0.009 despite inclusion of historical and post-discharge data. Performance of the LSTM model without historical data (*IP*) was also poor with ROC AUC of 0.516 ± 0.044. Inclusion of 30 days of historical data ($HIP_{30}$) resulted in a statistically significant improvement over both the baseline random forest and *IP* models, with an ROC AUC of 0.717 ± 0.011 (p<0.001). This indicates that *temporal* inclusion of historical data is important. Inclusion of 7 days of post-discharge context ($CHIP_7$) resulted in a small but statistically significant increase in performance, with ROC AUC of 0.727 ± 0.011 (p<0.05). Inclusion of 14 and 21 days of post-discharge context ($CHIP_{14}$ and $CHIP_{21}$) also resulted in slight but statistically significant improvements, with ROC AUC 0.740 ± 0.007 (p<0.001) and ROC AUC 0.763 ± 0.011 (p<0.001), respectively (Table 3). Precision-recall curves at varying operating points are plotted for $HIP_{30}$ and $CHIP_{21}$ models in Figure 3, demonstrating that post-discharge information improves model precision as the operating point is raised, whereas precision plateaus at around 0.8 without this context in $HIP_{30}$.

Because we did not pre-filter data to include only certain diagnoses or conditions, the patient population of this data is extremely diverse. Model attention was explored for three features due to their broad diagnostic and therapeutic relevance: Diagnosis Related Group (DRG), Primary Diagnosis (PDX), and Therapy Class (THERCLS). In order to exclude very rare features that would only be relevant in a small patient population, we only examined features that were present in at least 1% of the positive or negative samples. Due to the heterogeneity of data, even such a low threshold excluded the vast majority of features.

Diagnosis related group (DRG) contains 754 codes that reflect the primary diagnostic category for the admission of each patient. The top five predictive features for the model were sepsis, psychosis, gastrointestinal disorders, heart failure and shock, and chemotherapy (Figure 4). All top DRG features were all more frequently present in patients readmitted within 30 days than those who were not, with the two largest frequency differences seen in patients with psychoses and alcohol use.

Drug therapy class (THERCLS) contains 274 categories which are assigned to medications received during admission; therefore, patients may have a dozen or more entries per admission. Therapy class represents the middle of the range of drug information granularity, which is ordered as follows: therapy group (30 categories) > therapy class (300 categories) > generic drug name > specific drug name. The top features in this category were hypotensive agents, anticonvulsants, and anticoagulants (Figure 5). Both antipsychotics and antidepressants had large disparities in frequency between the two groups, indicating that psychiatric disorders are important factors for readmission. Non-steroidal anti-inflammatories (NSAIDs) were the fourth most important therapy class although the reasons remain unclear since NSAIDs are a common analgesic and not typically associated with high morbidity.

Finally, principal diagnosis (PDX) contains International Classification of Diseases, Tenth Revision (ICD-10) codes that represent the primary reason for admission with over 69,000 possible values. The large number of features results in low prevalence for any individual feature. The top feature for PDX was atheroembolism of the lower extremity which is typically associated with peripheral vascular disease which has high morbidity and surgical rates[38]. The second and third features were spondylosis and chondromalacia which are not typically associated with high morbidity, but are very common in elderly patients (Figure 6).

**Discussion**:

Excessive readmissions are an economic burden for hospitals, and a humanistic failure from the perspective of the healthcare team. For the average hospital, avoiding one excess readmission increases Medicare discharge reimbursement gains by up to $58,000[39]. Collectively, Readmissions cost Medicare up to $26 billion annually, $17 billion of which are potentially avoidable[4]. This cost is disproportionately distributed with larger hospitals in under resourced areas bearing the brunt of the cost and thus exacerbating healthcare inequalities between resource rich and economically oppressed populations[40,41]. As such, attempts to reduce excessive readmissions is a worthwhile endeavor from an economic, and humanitarian perspective.

Our work demonstrates the first time that *sequential* data has been utilized for re-admission modeling. This more closely reflects how a healthcare provider may assess the risk of patient re-admission, considering the sequence of recent events during and prior to an admission. The LSTM model significantly outperformed a baseline random forest model that does not consider time-series data. Similarly, the inclusion of sequential historical data prior to the admission substantially increased LSTM model performance compared to utilizing inpatient data alone. Lastly, we find that incorporating 7, 14, and 21 days of post-discharge data results in small, incremental benefits to model performance. This is an expected since follow up outpatient or pharmacy visits after discharge can provide insights into the patient's health status or medication compliance which will directly affect likelihood of readmission,

At a sensitivity of 80%, the *HIP* model is able to achieve approximately 60% specificity using inpatient and historical data. While these results are modest from a machine learning perspective, they are still clinically relevant considering the model is generating predictions for *all* hospitalized patients rather than specific disease conditions as is done in most prior work[11,41,42]. In addition, because the cost and morbidity associated with re-admission is significantly disproportionate to outpatient care[43], identifying and potentially intervening in at-risk patient represents a significant benefit[44] even if some outpatient resources are utilized for patients who are at lower risk.

Results from model attention were mixed, largely due to the complexity and very large feature space of the input data. Both Diagnosis Related Group and Therapy Class resulted in mostly reasonable features, however Principal Diagnosis results were less understandable Part of this difficulty may be that a feature prevalence of at least 1% was used as a cutoff when investigating the model attention results which could result in highly important but less prevalent features being excluded from our analysis. However, the model does consider *all* features during prediction meaning these features are still included. Overall, we show that patients both with severe acute conditions (e.g. sepsis, intracranial hemorrhage) or long-term chronic conditions (e.g. heart failure, chemotherapy, and digestive disorders) are at risk for 30-day readmission. For medications, cardiac, anti-infectives, and psychiatric drugs were associated with readmission. This is consistent with previous studies[45,46].

Because the model largely relies on structured data prior to and during hospitalization, we envision that it can be adapted in the hospital setting with direct input from the EHR. We will attempt this at our institution by leveraging existing EHR and imaging extraction pipeline[47]. We will also investigate the use of transformer networks [48] which have shown success in representing sequential data in language and may also be appliable to our work.

This study has some limitations. Reliance in insurance billing data means that some data may not be available at the time of discharge, however we believe the model could be adapted to receive HER input directly. Model attention results were also mixed, which could limit the end-user's ability to understand the model's prediction. Lastly, overall model performance was modest given that no pre-selection of patients was performed. Independent models could be created based on a patient's primary diagnosis, however this would decrease model generalizability and increase complexity for implementation.

**Authors' Contributions**

SK, JB, IB and HT designed and contributed to model development. FM and AA contributed to manuscript writing and editing. RD, MH, and JG contributed to statistical analysis and writing. All authors reviewed the final manuscript.

| Demographics | Diagnostic Data | Procedures | Payment | Medication |
|---|---|---|---|---|
| Age | Diagnosis at admission | Principle procedure during admission | Type of insurance plan | Drug name |
| Sex | Diagnoses 2-4 | Procedure performed | Co-insurance ($) | Drug generic name |
| Region | Diagnosis Related Group (DRG) | Number of invasive imaging studies | Copay ($) | Drug class |
| State | ER diagnosis | Number of total imaging studies | Deductible ($) | Drug group |
| Employee's Geographic Location | Emergency level (1-5) | | Sum of what patient paid (copay + deductible + coinsurance) ($) | Long term or short term drug |
| Employee's industry type | Injury site | | Average wholesale price of drug ($) | Number of days of medication |
| | Injury type | | Was provider in network? | Number of refills |
| | Trauma-related visit? | | | |

**Table 1**: Sample of included data elements from the IBM Marketscan Database roughly grouped by category. A full list of 200+ features used to train the model is available in Appendix A.

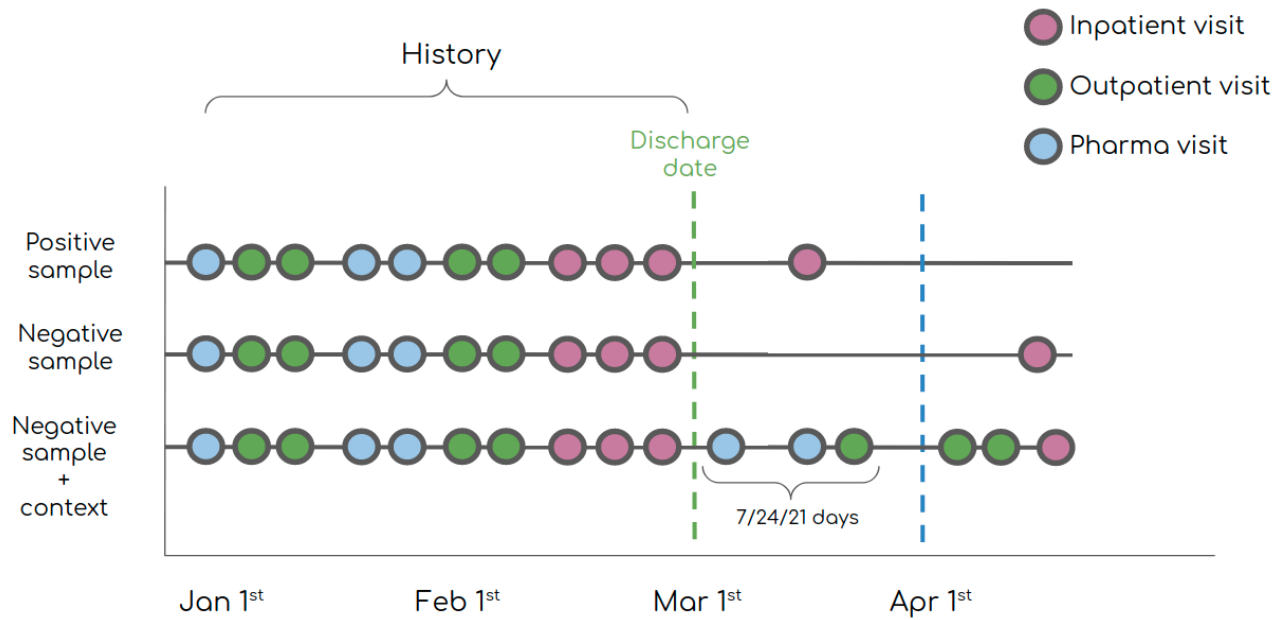

**Figure 1**: Illustration of positive and negative sample extraction. The major components for samples from left to right are i) 30-days historical data, ii) inpatient admission data, and iii) 7, 14, or 21 days of context. Positive samples are those in which a hospital discharge is followed by a re-admission within 30 days (top row). Negative samples are those in which a hospital discharge is not followed by a re-admission within 30 days, or the patient is never re-admitted (middle row). Individual models were trained using inpatient only data (*IP*), inpatient + historical data (*HIP*), or Inpatient + historical + 7, 14, or 21 days of context following discharge (*CHIP$_7$*, *CHIP$_{14}$*, or *CHIP$_{21}$*, last row).

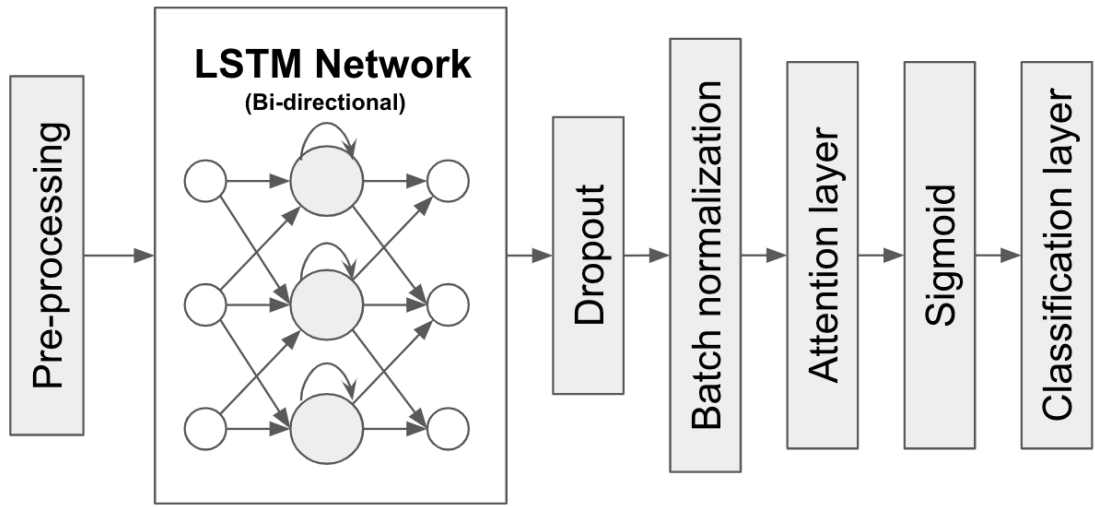

**Figure 2**. Design of the stacked bi-directional LSTM network with attention for feature probing. The predictive task was binary – return for an in-patient visit within 30 days of discharge (1) or not (0). The attention layer provides interpretable instance-level representations that may be used to probe the model. A Sigmoid activation layer yielded the highest performance.

|  | Readmitted in 30 days (n= 19,399) | Not re-admitted in 30 days (n= 19,399) |
| --- | --- | --- |
| Age | 49 ± 16 | 45 ± 17 |
| Male | 10366 | 10194 |
| Female | 9033 | 9205 |

Table 2. Patient characteristics for the positive and negative sequences. Note that no race information is provided in the MarketScan database to allow for subpopulation analysis. Negative sequences were undersampled to balance the training data.

| Model | Accuracy | Recall | Precision | F1-score | ROCAUC |
|---|---|---|---|---|---|
| Random Forest Classifier | 0.729 ± 0.016 | 0.440 ±0.019 | 0.400 ± 0.010 | 0.421 ± 0.010 | 0.510 ± 0.009 |
| IP | 0.732 ± 0.071 | 0.531 ± 0.042 | 0.652 ± 0.042 | 0.504 ± 0.031 | 0.516 ± 0.044 |
| $HIP_{30}$ | 0.663 ± 0.016 | 0.704 ± 0.034 | 0.655 ± 0.016 | 0.678 ± 0.014 | 0.717 ± 0.011 |
| $CHIP_7$ | 0.664 ± 0.021 | 0.746 ± 0.035 | 0.650 ± 0.006 | 0.694 ± 0.017 | 0.727 ± 0.007 |
| $CHIP_{14}$ | 0.667 ± 0.027 | 0.758 ± 0.023 | 0.657 ± 0.011 | 0.704 ± 0.007 | 0.740 ± 0.007 |
| $CHIP_{21}$ | 0.688 ± 0.017 | 0.722 ± 0.035 | 0.685 ± 0.011 | 0.702 ± 0.014 | 0.763 ± 0.011 |

**Table 3.** Performance metrics for all tested machine learning models. Baseline random forest classifier performance was poor, indicating that longitudinal representation of events is important. LSTM model performance using only inpatient data (*IP*) was also poor, suggesting that inpatient readmission cannot be predicted using only data from the inpatient visit. Model performance improved substantially when incorporating 30 days of historical data prior to admission (*HIP*), indicating that prior medical history is important. Finally, the model yielded incrementally improved results when incorporating 7, 14, and 21 days of additional data *after* discharge ($CHIP_7$, $CHIP_{14}$, and $CHIP_{21}$). Standard deviations shown across five folds of training/validation.

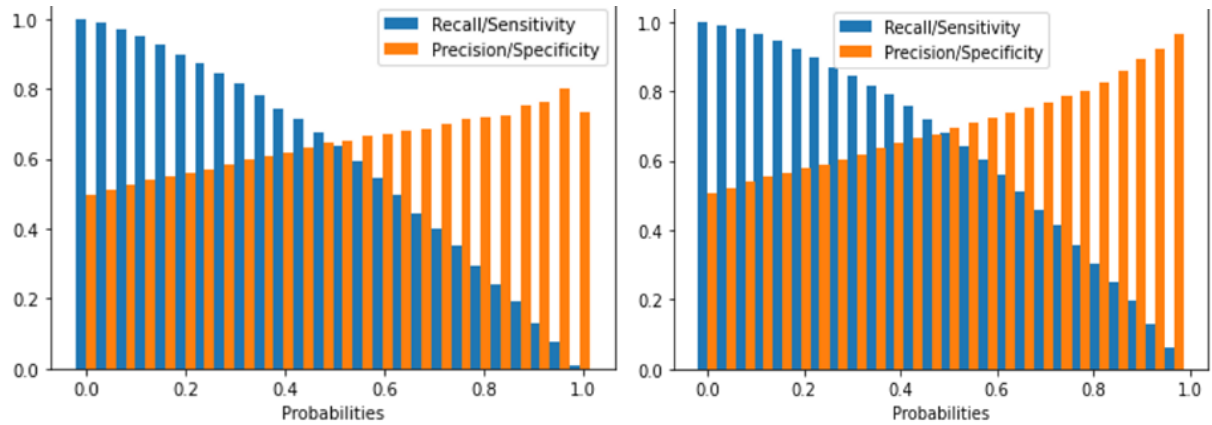

**Figure 3**. Recall/precision graphs for predicting in-patient re-admission. *Left*: performance for the *HIP* model which uses inpatient and historical data. *Right*: The top-performing *CHIP*$_{21}$ model that incorporates an additional 21 days of context following discharge. The *CHIP*$_{21}$ model demonstrates performance difference is noted as increased precision as the model threshold is increased for the top-performing model.

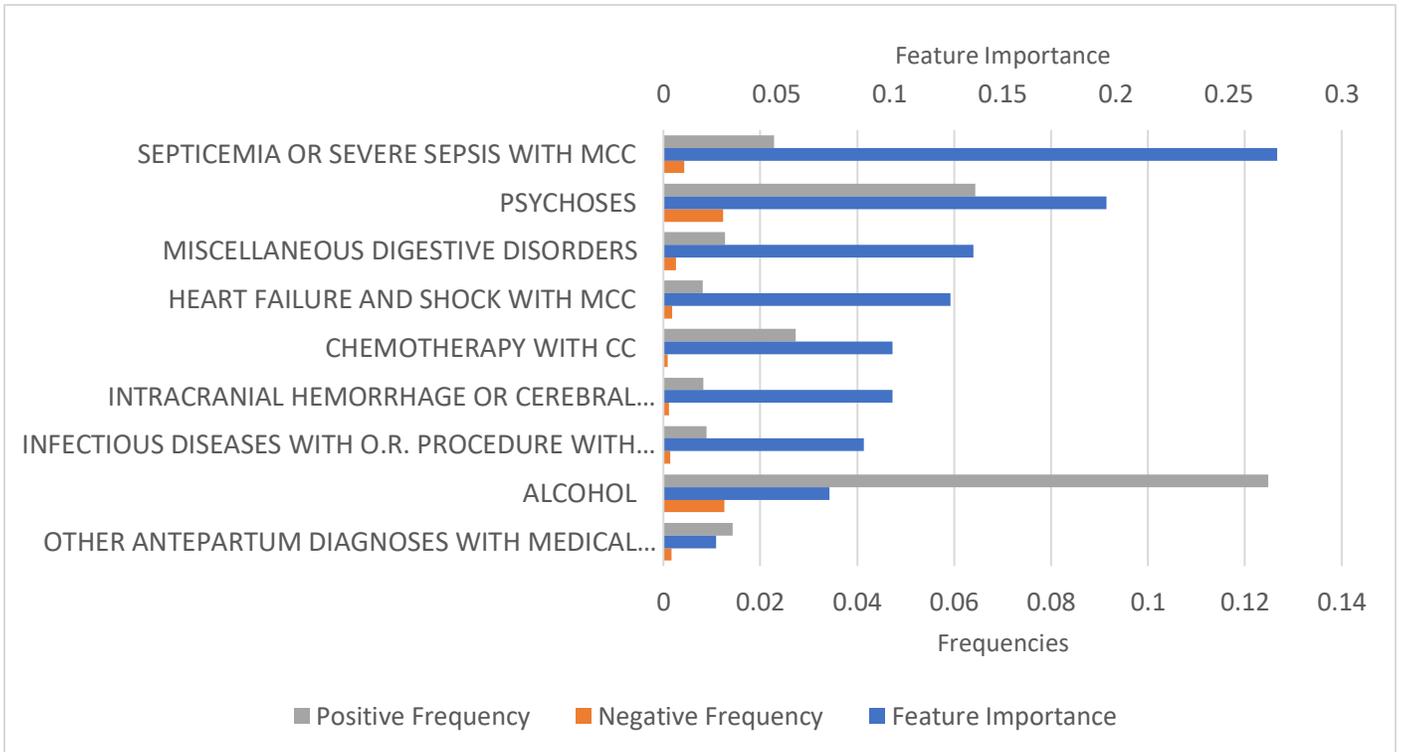

**Figure 4.** Top feature importance for diagnosis related group (DRG) that contributed to hospital re-admission. The top category was related to sepsis, followed by psychosis and digestive disorders. All features were present more frequently in patients who were readmitted CC = Complication or Comorbidity. MCC = Major Complication or Comorbidity.

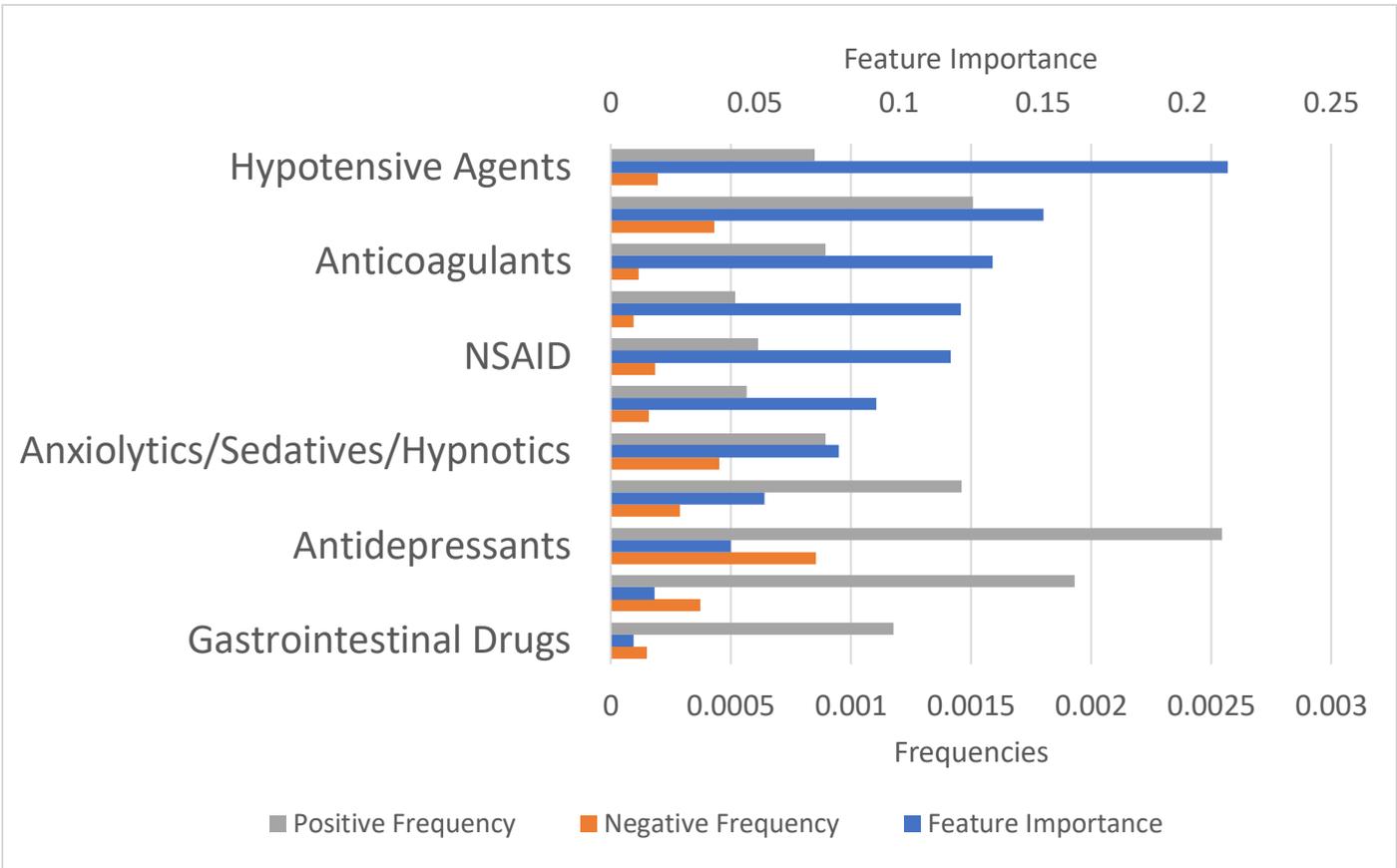

**Figure 5**. Top feature importance for Therapy Classes (THERCLS) that contributed to hospital re-admission. The top categories were hypotensive agents, anticonvulsants and anticoagulants which is reasonable given their high association with shock, seizure, thrombi, and surgery. All features were more common in patients who were readmitted. NSAID = Non-Steroid Anti-Inflammatory Drug

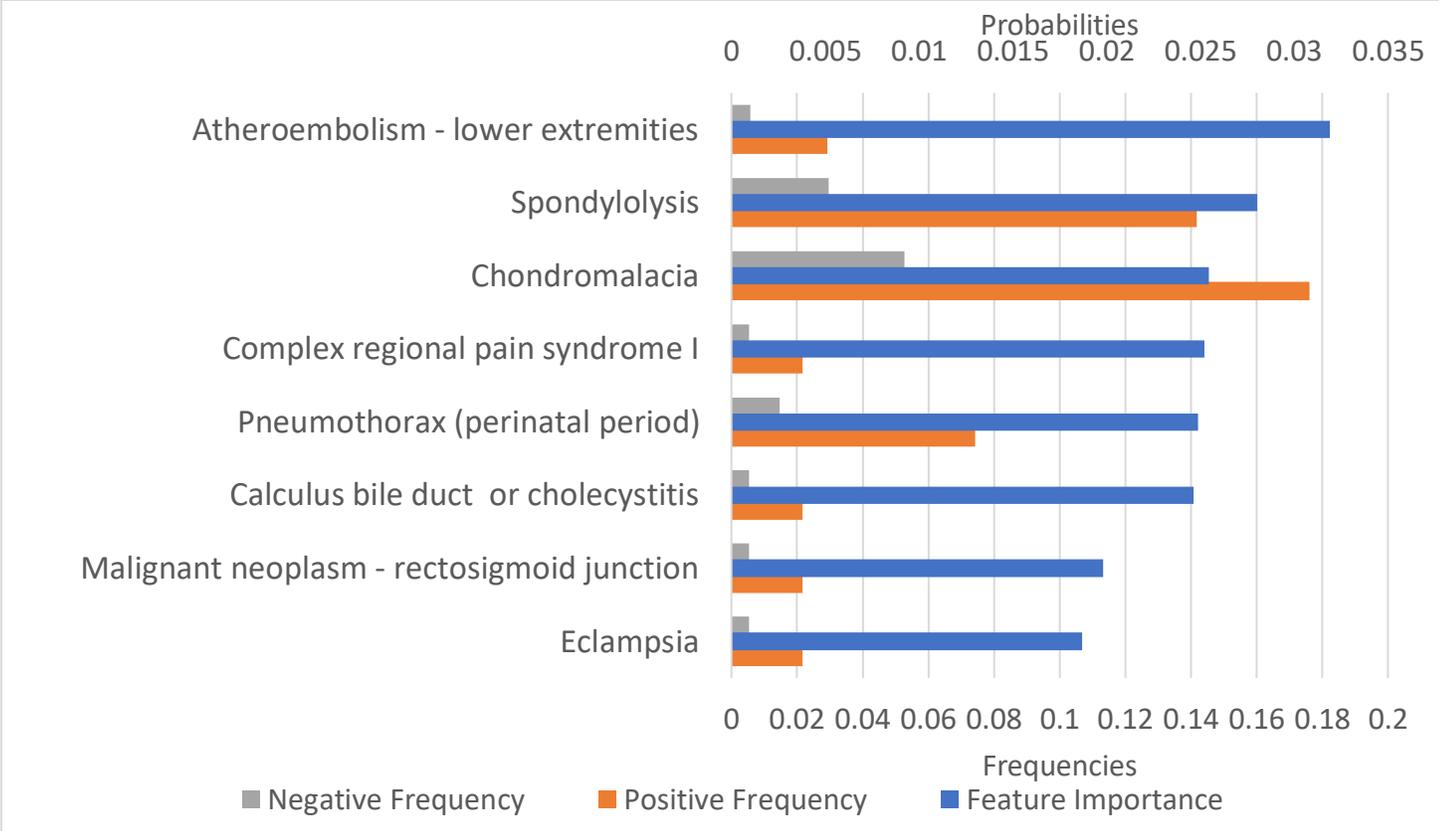

**Figure 6**. Top feature importances for Principal Diagnosis (PDX) that contributed to hospital re-admission. Lower extremity atheroembolism is reasonable considering the high morbidity and chronicity of peripheral vascular disease, however others like spondylosis and chondromalacia are more difficult to interpret as they are chronic conditions not independently associated with hospital admission.

**Appendix**

| Field Name | Description |
|---|---|
| ADMDATE | date of admission |
| ADMTYP | admission type |
| AGE | patient age |
| AGEGRP | age group - higher order variable |
| AWP | average wholesale price of drug |
| CASEID | identifier linking cases and services |
| CLSUPER | defined as people who have >4 visits per year; derived from N_Ervisits; "Clinical Superuser" |
| COINS | co-insurance |
| COPAY | copay |
| DAYSUPP | how many days supply of medication |
| DEACLAS | DEA class for controlled substances; |
| DEDUCT | deductible |
| DISDATE | date of hospital discharge (can be used to calculate length of stay in conjuction with date of admission) |
| DISPFEE | fee to dispense drug - combine with INGCOST for total cost of drug |
| DRG | daignosis related group - higher order diagnosis category |
| DX1 | primary diagnosis at admission |
| DX2 | secondary diagnosis at admission (can we get present on admission dx - POADx?) |
| DX3 | tertiary diagnosis at admission |
| DX4 | quaternary diagnosis at admission |
| DXVER | diagnosis version: 0 = ICD10, 9 = ICD9; keep only 0 |
| EESTATU | employee status - full time, part time, retired, COBRA |
| EESTATU | employment status (fulltime, partime, etc) |
| EGEOLOC | employee geographic location |
| EMDX | ER diagnosis |
| EMLEV | Emergency level (1-5) |
| ENROLID | enrollee ID |
| EPISID | unique identifer per row |
| ERDATE | visit date? |
| EROOP | sum of annual out of pocket payments for all ER services; what patient paid (copay + deductible + coinsurance) |
| ERPAY | sum of all payment to provider (patient + insurance); all doctors that provided service, not just ER doctors |
| FACHDID | facility ID |
| FACID | Facility ID? |
| FACIDIND | Unsure |
| FACNTWK | Unsure |
| FACPROF | facility vs professional claim; F=facility; P=professional |
| GENERID | generic drug ID - higher order variable |
| GENIND | indicates if drug has only one brand name, multiple brand names, one generic, or multiple generics |
| HCSID | Unsure |
| HINI | analog of h-index for researcher; At least 'h' publications have 'h' citations; this is based on #visits and #imaging during each; |
| HIRVU | similar to HINI but Total RVUs per visit, index over the year |
| IMSUPER | ED imaging superuser - definition used in Tarek's paper; defined as >10 imaging procedure in ED per year |
| INDSTRY | type of industry the employee works in |
| INGCOST | cost of drug itself; add to dispensing fee for total cost |
| INJURSITE1 | Injury site |
| INJURSITE2 | Injury site |
| INJURSITE3 | Injury site |
| INJURTYP | Injury type |
| MAINTIN | is this a long term or short term drug |
| MALE | |
| MDC | major diagnostic category |
| METQTY | number of tabs |
| MSA | metropolitan statistical area - what major metro area are they from? Rural = 0 |
| MSCLMID | unique per claim identifier; can use to aggregate all data from a claim |
| N_ERIMAGING | aggregrate per year, #imaging procedures obtained in that year |
| N_ERVISITS | per year |
| N_ERVISITS1IM | # er visits with at least one imaging procedure |
| N_Imaging | Number of total imaging studies |
| N_Invasive | Number of invasive imaging studies |
| NDCNUM | national drug code |

| NETPAY | net pay after coinsurance, deductible, etc |
|---|---|
| NPI | national provider identifier - encrypted |
| NTWKPROV | was provider in network |
| OOP | out of pocket cost |
| OOP_Imaging | out of pocket imaging cost |
| PAIDNTWK | was claim paid in network |
| PAY | gross pay before insurance |
| Pay_Imaging | gross pay for imaging? |
| PDX | priniciple diagnosis for admission |
| PHARMID | pharmacy ID |
| PLANTYP | type of plan: 1: Basic/major medical 2: Comprehensive<br>3: EPO<br>4: HMO<br>5: POS<br>6: PPO<br>7: POS with capitation 8: CDHP<br>9: HDHP |
| PLANTYP | Insurance plan type ( 0 means missing);  do not have non-insurance patients |
| PPROC | principle procedure during admission |
| PROC1 | Procedure performed |
| PROCGRP | group type of outpatient claim - higher lever variable |
| PROCMOD | procedure modifier prefix |
| PROCTYP | type of procedure code; 1 = CPT code; 7 = HCPC; 8 = CDT (ADA); keep only type 1 |
| PRVUFAC | Practive RVUs billed |
| REFILL | may indicate drug for acute illness versus maintenance drug |
| REGION | geographic region 1: Northeast<br>2: North Central 3: South<br>4: West<br>5: Unknown |
| REVCODE | Revenue code - type of  care received |
| RXMR | indicites of rx was filled by mail |
| SEX | patient gender |
| SINCELAST | Days since last ED visit |
| STATE | State |
| STDPLAC | place of service; 11 = office; 13 = assisted living; 22 = outpatient hospital; 23 = ER; 81= lab; 21 = inpatient |

| STDPROV | provider type; 220 = ER doctor; 500s are surgeons; 820 is NP; 845 is PA; 20-23 are mental health related; 38 hopsice; 140 pain mgmt; 202 osteopathic; 265 critical care |
|---|---|
| SVCDATE | date of service |
| SVCSCAT | service category |
| THERCLS | class of drug - higher order variable |
| THERGRP | higher order variable of THERCLS |
| TRAUMA | Visit related to trauma? |
| TRVUFAC | Total RVUs = radiologist work + practice expense + malpractice expense<br>This is the total resources involved in the study |
| UNITS | units of service |
| UNTILNEXT | days until next visit |
| WEEKDAY | Visit on a weekday? |
| WRVU | Work RVUs for imaging only |
| YEAR | Year |

**Appendix A**. Full list of data elements from the IBM Marketscan Dataset that were included in model development

| Hyperparameter | Value |
|---|---|
| Hidden layers | 150 |
| Stacked layers | 2 |
| Seq. length | 50 |
| Learning rate | 1e-5 |

**Appendix B.** Hyperparameters for the Bidirectional Long Short-Term Memory (LSTM) model